# Computing Cliques and Cavities in Networks

**Dinghua Shi[1]\*, Zhifeng Chen[2], Xiang Sun[2], Qinghua Chen[2], Chuang Ma[3], Yang Lou[4], Guanrong Chen[5]\***

[1]Department of Mathematics, College of Science, Shanghai University, Shanghai, China, shidh2012@sina.com
[2]Department of Statistics, School of Mathematics and Statistics, Fujian Normal University, Fuzhou, China
[3]Department of Internet Finance, School of Internet, Anhui University, Hefei, China
[4]Department of Computing and Decision Sciences, Lingnan University of Hong Kong, Hong Kong, China
[5]Department of Electrical Engineering, City University of Hong Kong, Hong Kong, China, eegchen@cityu.edu.hk

**Abstract:** Complex networks contain complete subgraphs such as nodes, edges, triangles, etc., referred to as simplices and cliques of different orders. Notably, cavities consisting of higher-order cliques play an important role in brain functions. Since searching for maximum cliques is an NP-complete problem, we use *k*-core decomposition to determine the computability of a given network. For a computable network, we design a search method with an implementable algorithm for finding cliques of different orders, obtaining also the Euler characteristic number. Then, we compute the Betti numbers by using the ranks of boundary matrices of adjacent cliques. Furthermore, we design an optimized algorithm for finding cavities of different orders. Finally, we apply the algorithm to the neuronal network of C. elegans with data from one typical dataset, and find all of its cliques and some cavities of different orders, providing a basis for further mathematical analysis and computation of its structure and function.

## Introduction

A network has three basic sub-structures: chain, star and cycle. Chains are closely related to the concept of average distance, while a small average distance and a large clustering coefficient together signifies a small-world network[1], where the clustering coefficient is determined by the number of triangles, which are special cycles. Stars follow some heterogeneous degree distribution, under which the growth of node numbers and a preferential attachment mechanism together distinguishes a scale-free network[2] from a random network[3]. Cycles not only contain triangles, but also have deeper and more complicated connotation. The cycle structure brings redundant paths into the network connectivity, creating feedback effects and higher-order interactions in network dynamics.

In [4], we introduced the notion of totally homogeneous networks in studying optimal network synchronization, which are networks with the same node degree, same girth (length of the smallest cycle passing the node in concern) and same path-sum (sum of all distances from other nodes to the node). We showed[4] that totally homogeneous networks are the ones of easily self-synchronizing among all networks of the same size. Recently, we identified[5] the special roles of two invariants of the network topology expressed by the numbers of cliques and cavities, the Euler characteristic number (alternative sum of the numbers of cliques of different orders) and the Betti number (number of cavities of different orders). In fact, higher-order cliques and smallest cavities are basic components of the totally homogenous networks.

More precisely, higher-order cycles of a connected undirected network include cliques and cavities. A clique is a fully-connected sub-network, e.g., a node is a 0-clique, an edge is a 1-clique, a triangle is a 2-cliques, and a complete graph of four nodes is a 3-clique, and so on, where the numbers indicate the orders. Also, a triangle is the smallest first-order cycle, which consists of three edges. The number of 1-cycles with different lengths (number of edges) is huge in a large-scale network. Similarly, a 3-clique is the smallest 2-cycle, and a 2-cycle contains some triangles. A chain is a broken cycle, where an edge is the shortest 1-chain, a triangle and two triangles adjacent by one edge are 2-chains, and so on, while a cycle is a closed chain. In the same manner, all these concepts can be extended to higher-order ones.

It is more challenging to study network cycles than node degrees; therefore, new mathematical concepts and tools are needed[6,7], including such as simplicial complex, boundary operator, cyclic operation and equivalent cycles, to classify various cycles and select their representatives for effective analysis and computation.

In higher-order topologies, the addition of two $k$-cycles, $c$ and $d$, is defined[5] by set operations as $c + d = (c \cup d) - (c \cap d)$. They are said to be equivalent, if $c + d$ is the boundary of a $(k + 1)$-chain[5]. All equivalent cycles constitute an equivalent class. Cavity is a cycle with the shortest length in each independent cycle-equivalent class (see Ref [5] for more details).

Cycles, cliques and cavities are found to play important roles in complex systems such as biological systems especially the brain. In the studies of the brain, computational neuroscience has a special focus on cyclic structures in neuronal networks. It was found[8] that cycles generate neural loops in the brain, which not only can transmit information all over the brain but also have an important feedback function. It was suggested[8] that such cyclic structures provide a foundation for the brain functions of memories and controls. Unlike cliques, which are placed at some particular locations (e.g. cerebral cortexes), cavities are distributed almost everywhere in the brain, connecting many different regions together. It is pointed out[9] that in both biological and artificial neural networks, one can find huge numbers of cliques and cavities which, being large and complex, have not been explored before. Of particular importance is that cavities play an indispensable role in brain functioning. All these findings indicate an encouraging and promising direction in brain science research. However, it remains unclear as how all such neuronal cliques and cavities are connected and organized. This calls for further endeavor into understanding the relationship between the complexity of higher-order topologies and the complexity of intrinsic neural functions in the brain. To do so, however, it is necessary to find most, if not all, cliques and especially cavities of different orders from the neuronal network.

Artificial intelligence, on the other hand, relies on artificial neural networks inspired by the brain neuronal network[10], including recurrent neural networks, convolutional neural networks, Hopfield neural networks, etc. Now, given the recent discovery of higher-order cliques and cavities in the brain, the question is how to further develop artificial intelligence to an even higher level by utilizing all the new knowledge about the brain topology. Notably, an effective neuronal network construction was recently proposed by a research team from the Massachusetts Institute of Technology, inspired by the real structure of neuronal network of the C. elegans[11].

It is important to understand how the brain stores information, learns new knowledge and reacts to external stimuli. It is also essential to understand how the brain adaptively creates topological connections and computing patterns. All these tasks depend on in-depth studies of the brain neuronal network. Recently, the Brain Initiative project of USA[12], the Human Brain project of EU[13] and the China Brain project[14] have been established to take such big challenges.

Many renowned mathematicians had contributed a lot of fundamental work to related subjects, such as Euler characteristic number, Betti number, groups of Abel and Galois, higher-order Laplacian matrices, as well as Euler-Poincaré formula and homology. This also demonstrates the importance of studying cliques and cavities for a further development of network science. In addition, the advance from pairwise interactions to higher-order interactions in complex system dynamics requires the knowledge of higher-order cliques and cavities of networks[15]. The numbers of zero eigenvalues of higher-order Hodge-Laplacian matrices are equal to the corresponding Betti numbers, while their associate eigenvectors are closely related to higher-order cavities[16].

Motivated by all the above observations, this paper studies the fundamental issue of computability of a complex network, based on which the investigation continues to find higher-order cliques and their Euler characteristic number, as well as all the Betti numbers and higher-order cavities. The proposed approach starts from $k$-core decomposition[17] and, through finding cliques of different orders, performs a sequence of computations on the ranks of the corresponding boundary matrices, to obtain all the Betti numbers. To that end, an optimized algorithm is developed for finding higher-order cavities. Finally, the optimized algorithm is applied to computing the neuronal network of C. elegans from a typical dataset, identifying its cliques and cavities of different orders.

**Results**

For computable undirected networks, the proposed approach is able to find all higher-order cliques, thereby obtaining the Euler characteristic number and all Betti numbers as well as some cavities of different orders. These can provide global information for understanding and analyzing the relationships between topologies and functions of various complex networks such as the brain neuronal network.

**1. Computable Networks**

For undirected networks, the cliques and their numbers of different orders in a network are both important sub-networks and parameters for analysis and computation. A simple example is shown in Figure 1, from which it is easy to find all cliques and their numbers of different orders.

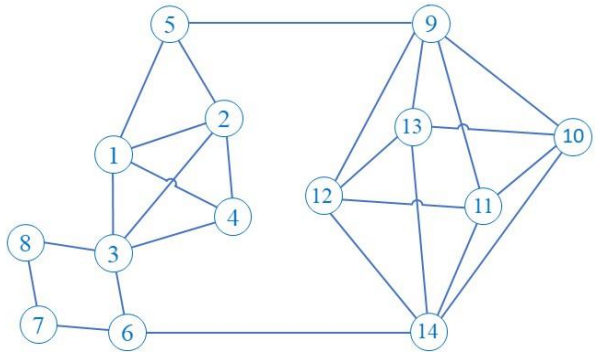

**Figure 1. A sample network.** This network has 14 nodes, 26 edges, 13 triangles, and 1 tetrahedron, where the numbers are node indexes (see Ref. [5]).

For a given general large-scale complex network, however, finding all cliques of different orders is never an easy task. In fact, even just searching for a maximum clique (namely, a clique with the largest possible number of nodes) from a large network is a computationally NP-complete problem[18]. It is well known that to find all cliques of a large-scale undirected network, especially when the network is dense, the number of cliques are huge and will increase exponentially as the network size becomes larger. For example, in the real USair, Jazz and Yeast networks[19], if the number of cliques is limited to not more than $10^7$ to be computable on a personal computer, the orders of the cliques are found to be only 9, 6 and 4, respectively, as summarized in Table 1. For these three real-world examples, it becomes impossible to compute the numbers of their higher-order cliques.

It is noticed that, even for large and dense networks, $k$-core decomposition can be used to efficiently determine their cells (layers), where the $k$-th cell has all nodes with degrees at least $k$, and the kernel of the network has the largest coreness value and is very dense. Therefore, the largest coreness value $k_{max}$ can be used to estimate the order of a maximum clique. For this reason, $k$-core decomposition is used to determine whether a given network is computable or not, subject to the available limited computing resources. If the computing resources allow the number of cliques, with the first several lowest orders, to be no more than $10^7$ to be computable, then the maximum coreness value should not be bigger than 30. In this paper, this coreness value threshold is set to $k_{max} = 25$, as detailed in Supplementary Note 1.

**2. Clique-Searching Method**

The Bron–Kerbosch algorithm[20] is a popular scheme for finding all cliques of an undirected graph, while the Hasse-diagram algorithm[9] is useful for finding all cliques of a directed network. For computable networks, we propose a method with an algorithm, named common-neighbors scheme, which can find all cliques of different orders and the associate Euler characteristic number.

In a network, the average degree is denoted by $\langle k \rangle$ and the number of edges by $|E|$. The computational complexity of the proposed algorithm is estimated to be $O(|E|\langle k \rangle)$ for finding all 2-cliques.

For illustration, the sample network shown in Figure 1 is used for clique searching, with a procedure in six steps, as follows:

(1) For each node, all its neighbors are listed, whose index numbers are bigger than the index number of this node:

Node 1 {2,3,4,5}, Node 2 {3,4,5}, Node 3 {4,6,8}, Node 4 {Ø}, Node 5 {9}, Node 6 {7,14}, Node 7 {8}, Node 8 {Ø}, Node 9 {10,11,12,13}, Node 10 {11,13,14}, Node 11 {12,14}, Node 12 {13,14}, Node 13 {14}, Node 14 {Ø}.

Then, the number of nodes in 0-clique is computed, yielding $m_0 = 14$.

(2) From the above list, edges are generated in increasing order of node indexes:

(1,2), (1,3), (1,4), (1,5), (2,3), (2,4), (2,5), (3,4), (3,6), (3,8), (5,9), (6,7), (6,14), (7,8), (9,10), (9,11), (9,12), (9,13), (10,11), (10,13), (10,14), (11,12), (11,14), (12,13), (12,14), (13,14).

Then, the number of edges in 1-clique is computed, yielding: $m_1 = 26$.

(3) For every edge, if its two end-nodes have a common neighbor whose index-number is bigger than the index-numbers of the two end-nodes, then all such neighbors are listed. For example:

edge (1,2) has common neighbors {3,4,5}, edge (1,3) has {4}, edge

(2,3) has {4}, edge (9,10) has {11,13}, edge (9,11) has {12}, edge (9,12) has {13}, edge (10,11) has {14}, edge (10,13) has {14}, edge (11,12) has {14}, edge (12,13) has {14}.

However, edge (1,4) and edges (1,5), (3,4), (3,6), (3,8), (5,9), (6,7), (6,14), (7,8), (9,13), (10,14), (11,14), (12,14), (13,14) do not have any common neighbor. Thus, the following triangles are obtained:

(1,2,3), (1,2,4), (1,2,5), (1,3,4), (2,3,4), (9,10,11), (9,10,13), (9,11,12), (9,12,13), (10,11,14), (10,13,14), (11,12,14), (12,13,14).

Then, the number of triangles in 2-cliques is computed, yielding: $m_2 = 13$.

(4) For each triangle, if its three nodes have a common neighbor whose index-number is bigger than the index-numbers of three nodes, then all such neighbors are listed.

Here, only the triangle (1,2,3) has a common neighbor, {4}, yielding 1 tetrahedron, (1,2,3,4).

Then, the number of tetrahedrons in 3-cliques is computed, yielding: $m_3 = 1$.

(5) The above procedure is continued, until it does not yield any more higher-order clique.

(6) Finally, the Euler characteristic number is computed, as follows[5]:

$$\chi = m_0 - m_1 + m_2 - m_3 = 14 - 26 + 13 - 1 = 0.$$

## 3. Computing Betti Numbers

The above-obtained cliques of various orders can be used to generate boundary matrices $B_k$, $k = 1, 2, \ldots$. Here, $B_1$ is the node-edge matrix, in which an element is 1 if the node is on the corresponding edge; otherwise, it is 0. Similarly, $B_2$ is the edge-face matrix, in which an element is 1 if the edge is on the corresponding face (triangle); otherwise, it is 0. All higher-order boundary matrices $B_k$ are generated in the same way. It is straightforward to compute the rank $r_k$ of matrices $B_k$ for $k = 1, 2, \ldots$, using linear row-column operations in the binary field, following the binary operation rules, namely $1 + 1 = 0$, $1 + 0 = 1$, $0 + 1 = 1$, $0 + 0 = 0$. Then, the Betti numbers[5] can be obtained by using formulas $\beta_k = m_k - r_k - r_{k+1}$, for $k = 1, 2, \ldots$.

One can also calculate the numbers of zero eigenvalues of higher-order Hodge-Laplacian matrices, so as to find the Betti numbers. To do so, some algebraic topology rules are needed to form oriented cliques[16].

As an example, the left-hand sub-network shown in Figure 1 is discussed, which has Nodes 1-8. The node-edge boundary matrix $B_1$ of rank $r_1 = 7$ is formed as follows, where the shaded row is linearly dependent on the other rows:

| $B_1$ | (1,2) | (1,3) | (1,4) | (1,5) | (2,3) | (2,4) | (2,5) | (3,4) | (3,6) | (3,8) | (6,7) | (7,8) |
|---|---|---|---|---|---|---|---|---|---|---|---|---|
| 1 | 1 | 1 | 1 | 1 | 0 | 0 | 0 | 0 | 0 | 0 | 0 | 0 |
| 2 | 1 | 0 | 0 | 0 | 1 | 1 | 1 | 0 | 0 | 0 | 0 | 0 |
| 3 | 0 | 1 | 0 | 0 | 1 | 0 | 0 | 1 | 1 | 1 | 0 | 0 |
| 4 | 0 | 0 | 1 | 0 | 0 | 1 | 0 | 1 | 0 | 0 | 0 | 0 |
| 5 | 0 | 0 | 0 | **1** | 0 | 0 | **1** | 0 | 0 | 0 | 0 | 0 |
| 6 | 0 | 0 | 0 | 0 | 0 | 0 | 0 | 0 | **1** | 0 | **1** | 0 |
| 7 | 0 | 0 | 0 | 0 | 0 | 0 | 0 | 0 | 0 | 0 | **1** | **1** |
| 8 | 0 | 0 | 0 | 0 | 0 | 0 | 0 | 0 | 0 | **1** | 0 | **1** |

Moreover, its edge-face boundary matrix of rank $r_2 = 4$ is obtained as follows, where the shaded column is linearly dependent on the other columns:

$$B_2 \quad \begin{array}{cccccc} & (1,2,3) & (1,2,4) & (1,2,5) & (1,3,4) & (2,3,4) \\ (1,2) & 1 & 1 & 1 & 0 & 0 \\ (1,3) & 1 & 0 & 0 & 1 & 0 \\ (1,4) & 0 & 1 & 0 & 1 & 0 \\ (1,5) & 0 & 0 & 1 & 0 & 0 \\ (2,3) & 1 & 0 & 0 & 0 & 1 \\ (2,4) & 0 & 1 & 0 & 0 & 1 \\ (2,5) & 0 & 0 & 1 & 0 & 0 \\ (3,4) & 0 & 0 & 0 & 1 & 1 \end{array}$$

Table 2 summarizes all calculation results for the sub-network on the left-hand side of the network shown in Figure 1, in which the Euler characteristic number and Betti numbers satisfy the Euler-Poincaré formula[5]

$$\chi = \beta_0 - \beta_1 + \beta_2 - \beta_3 = 1 + 2 - 1 - 0 = 0$$

**4. Cavity-Searching Method**

The concept of cavity comes from the homology group in algebraic topology. Since a large-scale network has many 1-cycles, for instance the network shown in Figure 1 has hundreds, to facilitate investigation they are classified into equivalent classes. In a network, each 1-cavity belongs to a linearly independent cycle-equivalent class[5] with the total number equal to the Betti number $\beta_1$. It is relatively easy to understand 1-cavity, which has boundary edges consisting of 1-cliques. Imagination is needed to understand higher-order cavities, which have boundary consisting of some higher-order cliques of the same order. In the literature, only one 2-cavity consisting of 8 triangles is found and reported[8]. In this paper, we found all possible smallest cavities and list them up to order 11 in Supplementary Note 2.

Since a cavity belongs to a cycle-equivalent class, only one representative from the class with the shortest length (namely, the smallest number of cliques) is chosen for further discussion. To find the smallest one, however, optimization is needed.

**4.1 Finding Cavity-Generating Cliques**

The procedure is as follows.

First, a maximum linearly independent group of column vectors is selected from the boundary matrix $B_k$, used as the minimum $k$th-order spanning tree, which consists of $r_k$ $k$-cliques, where $r_k$ is the rank value of the boundary matrix $B_k$ discussed above. Then, linear row-column binary operations are performed to reduce it to be in a simplest form. In every row of the resultant matrix, the column index of the first nonzero element is used as the index of the $k$-clique in the spanning tree. As an example, for the sub-network with Nodes 1-8 on the left-hand side of the network shown in Figure 1, the bold-faced 1's in matrix $B_1$ correspond to the columns indicated by (1, 2), (1, 3), (1, 4), (1, 5), (3, 6), (3, 8), (6, 7) shown at the top of the matrix, which constitute a spanning tree in the sub-network with Nodes 1-8 in Figure 1. It should be noted that the minimum $k$th-order spanning trees are not unique in general.

Next, the maximum group of linearly independent column vectors from boundary matrix $B_{k+1}$ is found, obtaining $r_{k+1}$ $(k+1)$-cliques as a group of linearly independent cliques. From this group, one continues to search for a $k$-clique that belongs to the boundary of the $(k+1)$-clique but does not belong to the $k$th-order spanning tree. In other words, the $r_{k+1}$ $k$-cliques should not be a $k$-clique in the minimum spanning tree. If this cannot be found, then one can choose another maximum group of linearly independent column vectors from boundary matrix $B_{k+1}$ and search again. In this way, $r_{k+1}$ $(k+1)$-cliques are found. As an example, for the sub-network with Nodes 1-8 in Figure 1, the bold-faced 1's in the boundary matrix $B_2$ correspond to the edges indicated by (2, 3), (2, 4), (2, 5), (3, 4). These are edges on the left-hand side of the boundary matrix $B_{k+1}$, which are different from the cliques in the spanning tree.

Then, the formula of Betti numbers, $\beta_k = m_k - r_k - r_{k+1}$, is used for computing, which is the number of linearly independent $k$-cavities. The task now is to find the rest $k$-cliques that are not in the $k$th-order minimum spanning tree and also not on the boundaries of linearly independent $(k+1)$-cliques. These are called cavity-generating cliques. In the same sub-network example, there is only one such clique: (7, 8). On the minimum spanning tree, after including all linearly independent boundaries, adding any cavity-generating $k$-clique will create a linearly independent $k$-cavity; in this example, the created one is the 1-cavity (3, 6, 7, 8).

**4.2 Searching Cavities by 0-1 Programming**

Every cavity-generating $k$-clique corresponds to at least one $k$-cavity. However, a cavity-generating $k$-clique may correspond to several cavities of different lengths, where the length is equal to the number of cliques. Since a cavity is a linearly independent cycle with the smallest number of cliques, the task of searching for a cavity can be reformulated as a 0-1 programming problem.

As shown above, there are $m_k$ $k$-cliques, $B_k$ is the boundary matrix between a $(k-1)$-clique and a $k$-clique, $B_{k+1}$ is the boundary matrix between a $k$-clique and a $(k+1)$-clique, and a $k$-cavity consists of some $k$-cliques. In the following, the vector space formed by $k$-cliques is denoted by $C_k$. A $k$-cavity can be expressed as $\mathbf{x} = (x_1, x_2, \ldots, x_{m_k}) \in C_k$, in which each component $x_i$ takes value 1 or 0, where 1 represents a $k$-clique with index $i$ in the cavity, while 0 means no such cliques there. Now, a cavity-generating $k$-clique with index $v$ is taken from all $k$-cliques and a vector $\mathbf{e} = (1, 1, \ldots, 1)^T$ is introduced. Then, the problem of searching for a $k$-cavity becomes the following optimization problem, which is to be solved for a nonzero solution:

$$\min_{\mathbf{x} \in C_k} f(\mathbf{x}) = \mathbf{x}\mathbf{e} \tag{1}$$
s.t. (i) $x_v = 1$,
(ii) $B_k \mathbf{x}^T = 0 \pmod 2$,
(iii) $\text{rank}(\mathbf{x}^T, B_{k+1})_{F_2} = 1 + r_{k+1}$.

Here, the first constraint means that the cavity comes from the cavity-generating $k$-clique with index $v$. The second constraint implies that the cavity is a $k$-cycle, namely the boundaries of $k$-cliques that form the cavity should appear in pairs. The third constraint shows that the $k$-cavity to be found will not be a linear representation of the $(k + 1)$-cliques, where $F_2$ indicates that the operations are performed in the binary field, which can avoid generating false cavities.

To ensure that the $\beta_k$ cavities found are linearly independent, the following 0-1 programming is performed, where $\mathbf{x}^{(l)}$ is the $l$-th $k$-cavity: (i) $x_v^{(l)} = 1$; (ii) $B_k \mathbf{x}^{(l)T} = 0 \pmod 2$; (iii) $\text{rank}(\mathbf{x}^{(1)T}, \cdots, \mathbf{x}^{(l)T}, B_{k+1})_{F_2} = l + r_{k+1}$, for $l = 1, 2, \ldots, \beta_k$.

It was found that the sample sub-network shown on the left-hand side of Figure 1 has two 1-cavities, where two cavity-generating 1-cliques are $x_{14} = 1$ corresponding to edge (7, 8) and $x_{11} = 1$ corresponding to edge (5, 9). This optimization is detailed as follows:

$$\min_{\mathbf{x} \in C_1} f(\mathbf{x}) = \mathbf{x}\mathbf{e} \tag{2}$$
s.t. (i) $x_{14} = 1$, (ii) $B_1 \mathbf{x}^T = 0 \pmod 2$, namely
$x_1+x_2+x_3+x_4=0$, $x_1+x_5+x_6+x_7=0$, $x_2+x_5+x_8+x_9+x_{10}=0$, $x_3+x_6+x_8=0$, $x_4+x_7+x_{11}=0$,
$x_9+x_{12}+x_{13}=0$, $x_{12}+x_{14}=0$, $x_{10}+x_{14}=0$, $x_{11}+x_{15}+x_{16}+x_{17}+x_{18}=0$, $x_{15}+x_{19}+x_{20}+x_{21}=0$,
$x_{16}+x_{19}+x_{22}+x_{23}=0$, $x_{17}+x_{22}+x_{24}+x_{25}=0$, $x_{18}+x_{20}+x_{24}+x_{26}=0$, $x_{13}+x_{21}+x_{23}+x_{25}+x_{26}=0$,
(iii) $\text{rank}(\mathbf{x}^T, B_2)_{F_2} = 1 + r_2$.

By solving the above 0-1 programming problem, with $x_{14} = 1$ corresponding to (7, 8), it yields $x_{10} = 1$ corresponding to (3, 8), and with $x_{12} = 1$ corresponding to (6, 7), it yields $x_9 = 1$ corresponding to (3, 6), which generates the first cavity (3, 6, 7, 8). Then, replacing $x_{14} = 1$ with $x_{11} = 1$ yields the second cavity (1, 5, 9, 10, 14, 6, 3), which has 8 equal-length cavities, including 1-cavity (2, 5, 9, 10, 14, 6, 3) and 1-cavity (1, 5, 9, 11, 14, 6, 3), etc. Finally, checking the $\text{rank}(\mathbf{x}^{(1)T}, \cdots, \mathbf{x}^{(l)T}, B_2)_{F_2} = l + r_2$, for $l = 1, 2$, verifies that the optimization meets all the constraints.

## 5. Cliques and Cavities of C. elegans

For a dataset of C. elegans with 297 neurons and 2148 synapses[21], all cliques and some cavities are obtained here by using the above-described techniques and algorithms, which is compared to the homogeneous Erdös-Rényi (ER) random network with the same numbers of nodes and edges. The results are shown in Figure 2 and Table 3.

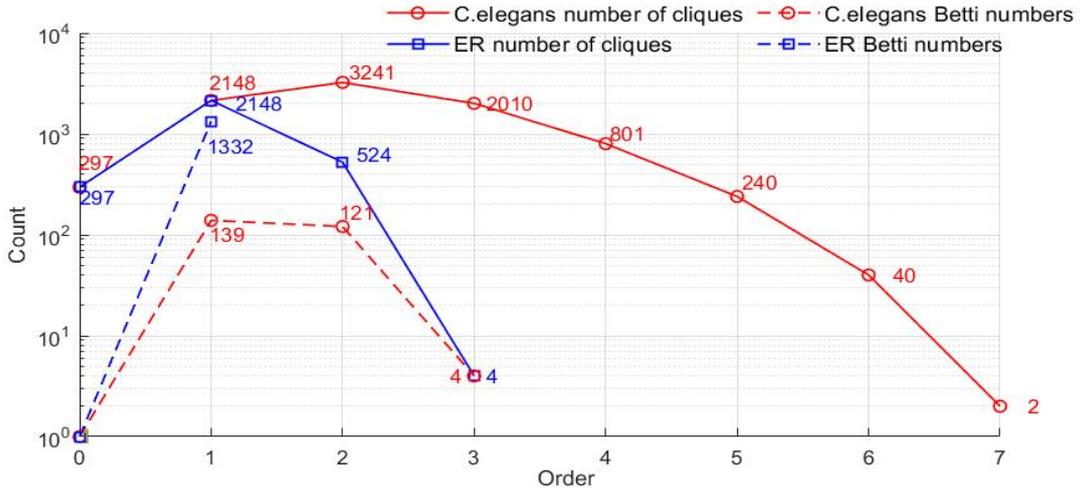

**Figure 2. Cliques and Betti numbers.** The number of cliques and the Betti numbers for the C. elegans neuronal network versus the Erdös-Rényi (ER) random network

Since the highest-order nonzero Betti number is $\beta_3 = 4$, the C. elegans has 4 linearly independent 3-cavities, and these 4 cavities have cavity-generating 3-cliques {164, 163, 119, 118}, {119, 167, 118, 227}, {195, 185, 119, 118} and {227, 195, 119, 118}. The cavity-generating 3-clique {164, 163, 119, 118} forms a 3-cavity with 8 nodes {85, 13, 3, 164, 163, 119, 118, 158}, which is the smallest 3-cavities[5], with structures shown in Figure 3 (a). The cavity-generating 3-clique {119, 167, 118, 227} forms a 3-cavity with 11 nodes {163, 3, 162, 119, 154, 167, 118, 227, 85, 13, 164} as shown in Figure 3 (b). The cavity-generating 3-clique {195, 185, 119, 118} forms a 3-cavity with 8 nodes {171, 13, 3, 195, 185, 119, 118, 173}, as shown in Figure 3 (c). The cavity-generating 3-clique {227, 195, 119, 118} forms a 3-cavity with 8 nodes {173, 13, 3, 227, 195, 119, 118, 185}, as shown in Figure 3 (d). All details are included in Supplementary Note 3.

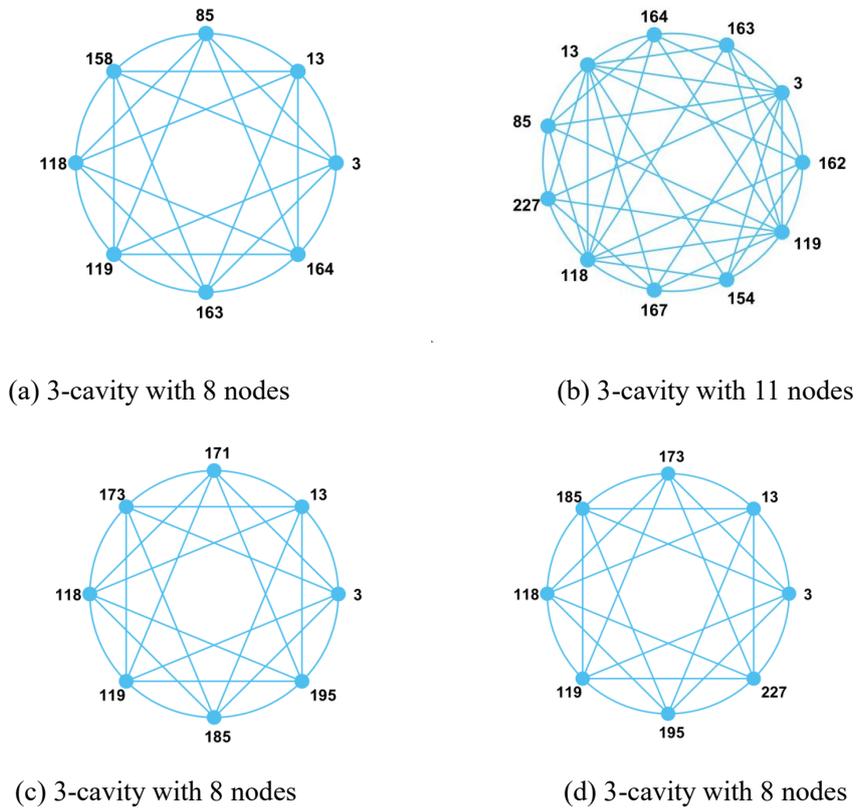

(a) 3-cavity with 8 nodes  (b) 3-cavity with 11 nodes

(c) 3-cavity with 8 nodes  (d) 3-cavity with 8 nodes

**Figure 3. Four 3-cavities in the C. elegans neuronal network.** (a) 3-cavity with 8 nodes; (b) 3-cavity with 11 nodes; (c) 3-cavity with 8 nodes; (d) 3-cavity with 8 nodes

**Discussions**

For a given directed network, how can one analyze its higher-order cliques and cavities? In [9], a Hasse algorithm was designed to find all directed cliques. However, both concepts of cycle and cavity were not precisely defined there. For an undirected network, the length of a cavity, namely the number of cliques that compose it, is longer than the lengths of the cliques (1-order higher cavity) as a cycle. For example, an undirected triangle of length 3 not only is a 2-clique but also is a 1-cycle, while 1-cavity at least is a quadrangle of length 4. For a directed network, however, this may not be true. For instance, the smallest directed 1-cavity could be composed by two reversely directed edges between two nodes, where both edges have length 2. But, a directed 2-clique could be a directed triangle of length 3. This shows the extreme complexity of directed cavities, which will be a research topic for future investigation.

It should be noted that the key technique used in this paper is to examine various combinations of cliques and cavities, which differs from the studies based on node degrees in the current literature, where the focus is on statistical rather than topological properties of the network. After comparing the neuronal network of a C. elegans to a random network, it was found that they are very different regarding the numbers of cliques and cavities. From the perspective of brain science, various combinations of higher-order topological components such as cliques and cavities are of extreme importance, without which it is very difficult or even impossible to understand and explain the functional complexity of the brain. In fact, this provides reasonable supports to many recent studies on the brain[12,13,14].

The intrinsic combination of cliques and cavities also brings some unexpected problems to programming the proposed optimization algorithm. For example, because a minimum spanning tree of a network is not unique, the algorithm may not produce the expected results when searching for cavities. Efforts have been made to determine the information of cavities by eigenvectors corresponding to zero eigenvalues of higher-order Hodge-Laplacian matrices. However, similar non-uniqueness problem occurred in finding eigenvectors, demonstrating the extreme complexity of the clique and cavity-searching problems.

**Method and Algorithms**

To solve the above optimization problem (Eq. (1)) with $\beta_k$ $k$-cavities is difficult due to the third constraint in the optimization, and because the minimum spanning tree is not unique. As a remedy, the optimization problem is separated into two parts.

The first part is to use the following 0-1 programming to search for all possible cycles that contain the $\beta_k$ $k$-cavities $x^{(l)}$, $l = 1, 2, \ldots, \beta_k$:

$$\min_{x \in C_k} f(x) = x^{(l)} e \qquad (3)$$

s.t. (i) $x_v^{(l)} = 1$;

(ii) $B_k x^{(l)T} = 0 \pmod{2}$

The second part is to use the third constraint to find all the $\beta_k$ $k$-cavities within all possible cycles, i.e. to check if $\mathrm{rank}(x^{(1)T}, \cdots, x^{(l)T}, B_{k+1})_{F_2} = l + r_{k+1}$, for the $l$-th cavity $x^{(l)}$, $l = 1, 2, \ldots, \beta_k$.

**1. Searching for specific cycles**

Because there is a constraint $B_k x^{(l)T} = 0 \pmod{2}$ in Eq. (3), it is not a traditional 0-1 linear programming problem. To reformulate the problem, a couple of notations are introduced: $\tilde{B}_k = [B_k, -2I]$ and $\tilde{x}^{(l)T} = [x^{(l)}, y]$, where $I$ is the identity matrix and $y = [y_1, \ldots, y_{m_k}]$. Then, $B_k x^{(l)T} = 0 \pmod{2}$ can be equivalently rewritten as $\tilde{B}_k \tilde{x}^{(l)T} = 0$. Since the minimum length of the $k$-cavity is $L_{\min} = 2^{k+1}$, Eq. (3) can be transformed to the following 0-1 programming problem for a linear system of equations:

$$\min_{x \in C_k} f(x) = x^{(l)}e \tag{4}$$

(i) $x_v^{(l)} = 1$,

(ii) $x^{(l)}e = L_{\min}$,

(iii) $\tilde{B}_k \tilde{x}^{(l)T} = 0$,

(iv) $x_i, y_i = 0$ or $1$, $l = 1, 2, \ldots, \beta_k$.

Equation (4) can be solved by using Matlab toolbox for 0-1 linear system of equations, and the algorithm is described as follows.

**Algorithm 1**: Searching specific cycles ($x^*$ = Find Cycle ($B_k$, $v$, $L_{\min}$))

Input: boundary matrix $B_k$

indices of all cavity-generating cliques $\{v^1, \ldots, v^{\beta_k}\}$

length of the smallest cycle $L_{\min} = 2^{k+1}$

Output: specific cycles $x^*$

## 2. Finding all cavities

The third constraint in the optimization problems (Eqs. (3) and (4)) is needed to check, so as to identify which cycles found by Algorithm 1 are $k$-cavities and then to determine their cavity-generating cliques. For cavity-generating cliques not included in the list in Algorithm 1, or if there are many cavity-generating cliques appearing in the same cavity, one has to search new cycles obtained by Algorithm 1 again by increasing the lengths of the $2^{k-1}$ cliques.

Summarizing the above steps gives the following cavity-searching algorithm:

**Algorithm 2**: Checking all $k$-cavities ($\{x_1^*, \ldots, x_{\beta_k}^*\}$ = Find Cavity ($B_k$, $B_{k+1}$, $v$, $L_{\min}$))

Input: boundary matrices $B_k$ and $B_{k+1}$

indices of all cavity-generating cliques $\{v^1, \ldots, v^{\beta_k}\}$

length of cycle $L_{\min} + j2^{k-1}$, $j = 0, 1, \ldots$

Output: all $k$-cavities $\{x_1^*, \ldots, x_{\beta_k}^*\}$

**Data availability**

Data used in this work can be accessed at http://linkprediction.org/index.php/link/resource/data/1

**Code availability**

The code for the numerical simulations presented in this article is available from the corresponding authors upon reasonable request.

# Acknowledgements

The authors would like to thank the research supports by the National Natural Science Foundation of China (Grants No. 61174160, 12005001), the National Natural Science Foundation of Fujian Province (Grant No. 2019J01427), the Program for Probability and Statistics: Theory and Application (No. IRTL 1704), and by the Hong Kong

Research Grants Council through General Research Funds (Grant CityU11206320).

**Author contributions**
DS and GC developed the theory and wrote the text. ZC, XS, CM, YL and QC performed the simulations and computations for cross check. All authors checked and verified the entire manuscript.

**Competing interests**
Authors declare no competing interests.

Correspondence and requests for materials should be addressed to DS and GC.

**Table 1.** Three real networks: their sizes (number of nodes |N| and number of edges |E|), maximum coreness $k_{max}$, maximum number of cliques $c_{max}$, number of $k$-cliques $m_k$, and the maximum order $k$ of cliques with their numbers $m_k < 10^7$

| Network | |N| | |E| | $k_{max}$ | $c_{max}$ | max$\{k|m_k <10^7\}$ |
|---|---|---|---|---|---|
| USAir | 332 | 2126 | 26 | 21 | 9  ($m_9$= 9121594) |
| Jazz | 198 | 2742 | 29 | 29 | 6  ($m_6$= 2416059) |
| Yeast | 2375 | 11693 | 40 | Unknown | 4  ($m_4$= 2454474) |

**Table 2.** Computational results for the sub-network on the left-hand side of the network shown in Figure 1. Here, $m_k$ is the number of $k$-cliques, $r_k$ is the rank of the boundary matrix $B_k$, and $\beta_k$ is the $k$th Betti number

| Order $k$ | 0 | 1 | 2 | 3 |
|---|---|---|---|---|
| $m_k$ | 14 | 26 | 13 | 1 |
| $r_k$ | 0 | 13 | 11 | 1 |
| $\beta_k = m_k - r_k - r_{k+1}$ | 1 | 2 | 1 | 0 |

**Table 3.** Euler characteristic number, Betti numbers and the Euler-Poincaré formula for the C. elegans network and the Erdös-Rényi (ER) network

| Network | The Euler characteristic number, Betti numbers and the Euler-Poincaré formula |
|---|---|
| C. elegans | $\chi$ = 297−2148+3241−2010+801−240+40−2 = 1−139+121−4 =  −21 |
| ER | $\chi$ = 297−2148+524−4 = 1−1332 =  −1331 |

# Computing Cliques and Cavities in Networks

**Dinghua Shi[1]\*, Zhifeng Chen[2], Xiang Sun[2], Qinghua Chen[2]\*, Chuang Ma[3], Yang Lou[4], Guanrong Chen[5]\***

[1]Department of Mathematics, College of Science, Shanghai University, Shanghai, China, shidh2012@sina.com
[2]Department of Statistics, School of Mathematics and Statistics, Fujian Normal University, Fuzhou, China
[3]Department of Internet Finance, School of Internet, Anhui University, Hefei, China
[4]Department of Computing and Decision Sciences, Lingnan University of Hong Kong, Hong Kong, China
[5]Department of Electrical Engineering, City University of Hong Kong, Hong Kong, China, eegchen@cityu.edu.hk

# Supplementary Information

**Supplementary Note 1. $k$-Cores and Computable Networks**

For the real USair, Jazz and Yeast networks[19], the number of cliques of different orders is limited to not more than $10^7$ as detailed in Table SI-1.

**Table SI-1** Number of cliques of different orders in three real networks

| Network | 0-cliques | 0-cliques | 2-cliques | 3-cliques | 4-cliques |
|---|---|---|---|---|---|
| USAir | 332 | 2126 | 12181 | 61072 | 243506 |
| Jazz | 198 | 2742 | 17899 | 78442 | 273697 |
| Yeast | 2375 | 11693 | 60689 | 424444 | **2454474** |

| Network | 5-cliques | 6-cliques | 7-cliques | 8-cliques | 9-cliques |
|---|---|---|---|---|---|
| USAir | 766659 | 1931547 | 3947163 | 6608097 | **9121594** |
| Jazz | 845960 | **2416059** | | | |
| Yeast | | | | | |

If the number of cliques is limited to not more than $10^7$ to be computable on a personal computer, the orders of the cliques are found to be only 9, 6 and 4, respectively, as indicated by the three bold numbers in Table 1. For these three real-world examples, it becomes impossible to compute the numbers of their higher-order cliques.

It is noted that in any network of a fixed size, except trees, its number of cliques of different orders has a peak value as the order number increases, namely it is first increasing and then decreasing. For instance, for a fully-connected network of size $N$, the numbers of its $m$-th order cliques are: $m_0 = C_m^1$, $m_1 = C_m^2$, $\cdots$, $m_{N-2} = C_N^{N-1}$, $m_{N-1} = C_N^N$, where it peaks at the $\left(\frac{N}{2} - 1\right)$-clique (if $N$ is even) or the $\left(\frac{N-1}{2} - 1\right)$-clique (if $N$ is odd). Specifically, when $N = 30$, it peaks at the 14-clique, with $m_{14}$ =155117520; when $N = 25$, it peaks at the 12-clique, with $m_{12} = C_{25}^{13} = 5,200,300$.

Given limited computational resources, how can one determine if a given network is computable? For relatively large-scale and dense networks, $k$-core decomposition[17] can provide an estimate. The $k$-core technique is used to determine the cells of different orders, where all nodes on the $k$-sell have degrees larger than or equal to $k$. The cell with the largest coreness value is the core of the network, in which the connection is dense; therefore, it can be used to measure the order of the largest clique in the network. For example, in the Jazz network, the 29th cell has 30 nodes and 435 edges, implying that this is a fully-connected network; therefore, its core is a 29-clique, which also shows the order of the largest clique in the Jazz network. In the USAir network, the largest coreness value is 26, where the core has 35 nodes and 539 edges; therefore, its largest clique is a 21-clique, which is close to the true coreness value 26. For the Yeast network, its core has 64 nodes and 1623 edges, which is known to have the largest coreness value 40; although the computation here reaches up to 6-cliques, it can be seen that the order of the largest clique would not be small. The detailed coreness values of USAir, Jazz and Yeast are summarized in Table SI-2, where $m_i$ is the coreness value of the $i$-core, $i = 0, 1, 2, ..., 29$.

**Table SI-2** Coreness values of three real networks

| Core value | $m_0$ | $m_1$ | $m_2$ | $m_3$ | $m_4$ | $m_5$ | $m_6$ | $m_7$ | $m_8$ | $m_9$ |
|---|---|---|---|---|---|---|---|---|---|---|
| USAir-26 | 35 | 539 | 4938 | 30580 | 137428 | 468604 | 1248988 | 2656044 | 4570650 | 6425067 |
| Jazz-29 | 30 | 435 | 4060 | 27405 | 142506 | 593775 | 2035800 | 5852925 | 14307175 | 30045015 |
| Yeast-40 | 64 | 1623 | 22344 | 196991 | 1222179 | 5656082 | 20278476 | | | |
| Core value | $m_{10}$ | $m_{11}$ | $m_{12}$ | $m_{13}$ | $m_{14}$ | $m_{15}$ | $m_{16}$ | $m_{17}$ | $m_{18}$ | $m_{19}$ |
| USAir-26 | 7419660 | 7055424 | 5520504 | 3540415 | 1847164 | 774823 | 256755 | 65498 | 12370 | 1624 |
| Jazz-29 | 54627300 | 86493225 | 119759850 | 145422675 | 155117520 | 145422675 | 119759850 | 86493225 | 54627300 | 30045015 |
| Yeast-40 | | | | | | | | | | |
| Core value | $m_{20}$ | $m_{21}$ | $m_{22}$ | $m_{23}$ | $m_{24}$ | $m_{25}$ | $m_{26}$ | $m_{27}$ | $m_{28}$ | $m_{29}$ |
| USAir-26 | 132 | 5 | 0 | | | | | | | |
| Jazz-29 | 14307175 | 5852925 | 2035800 | 593775 | 142506 | 27405 | 4060 | 435 | 30 | 1 |
| Yeast-40 | | | | | | | | | | |

The above analysis shows that, given the limited computational resources today, if the number of $k$-cliques is up to the order of $10^7$ then the largest coreness value of a network should not be larger than 30, or even should not be larger than 25 to be feasible on a personal computer.

If the $k$-core decomposition is performed by removing all nodes of degree $k = 1$, then some new nodes of degree $k \leq 1$ may emerge, and these nodes have to be removed as well, until all nodes have degree $k > 1$. All removed nodes and edges constitute 1-core with coreness value 1. This process continues for $k = 2, 3, \ldots$, until the highest value $k_{max}$ at which all nodes will be removed, and this last core has a coreness value $k_{max}$, which is the core of the original network.

The same idea can be applied to compute cliques, which is named $k$-clique decomposition. For illustration, the sample network shown in Figure 1 is discussed again. This network does not have 0-core and 1-core, and its 2-core contains nodes 6, 7, 8 and edges (3,6), (3,8), (6,7), (6,14), (7,8). Its 1-clique is composed of edges (3,6), (3,8), (6,7), (6,14), (7,8). Its 3rd sell consists of nodes 1, 2, 3, 4, 5 and edges (1,2), (1,3), (1,4), (1,5), (2,3), (2,4), (2,5), (3,4), (5,9). Its 2-clique is composed of edges (1,2,5), (9,10,11), (9,10,13), (9,11,12), (9,12,13), (10,11,14), (10,13,14), (11,12,14), (12,13,14). Its 4th cell consists of nodes 9, 10, 11, 12, 13, 14 and edges (9,10), (9,11), (9,12), (9,13), (10,11), (10,13), (10,14), (11,12), (11,14), (12,13), (12,14), (13,14). Its 3-cliques is composed of (1, 2, 3, 4). This example shows the difference between the $k$-coe decomposition and the $k$-clique decomposition.

**Supplementary Note 2. Smallest Possible Cavities of Different Orders**

The concept of cavity comes from homology group in algebraic topology. Cavity is a special topological structure. The 1-cavity and 2-cavity were found from observation[8]. In general, a smallest $n$-cavity is a smallest cycle consisting of some $n$-cliques, where the number of such $n$-cliques is larger than the number of the boundaries of $(n + 1)$-cliques. Furthermore, a smallest $n$-cavity can be obtained by introducing 2 more nodes, each connecting to all nodes in the smallest $(n - 1)$-cavity. Today, it is suspected that there is as high as 11th-order cavity in the neural network of the brain[9]. It is also known that the smallest $k$-cavity has a characteristic number[5] $\chi = 1 + (-1)^k$.

Numbers and features of smallest cavities of order 1 to order 11 are summarized in Figure SI-1.

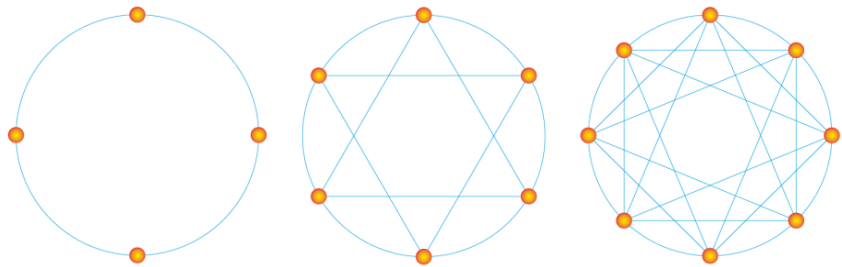

1-cavity: $m_0$=4，$m_1$=4； $\chi$=0

2-cavity: $m_0=6$, $m_1=12$, $m_2=8$; $\chi=2$

3-cavity: $m_0=8$, $m_1=24$, $m_2=32$, $m_3=16$; $\chi=0$

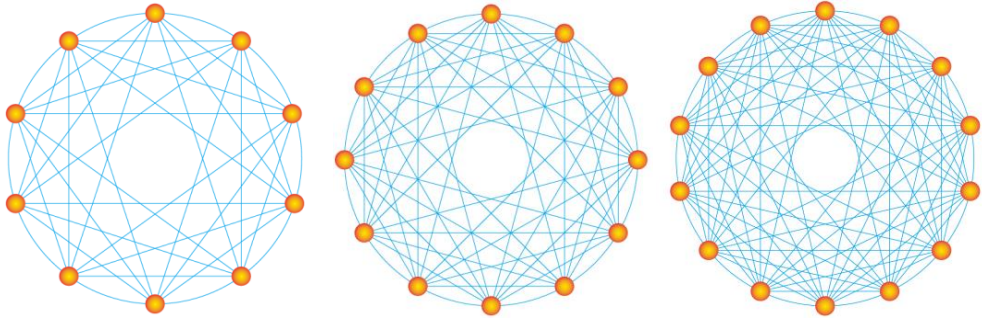

4-cavity: $m_0=10$, $m_1=40$, $m_2=40$, $m_3=80$, $m_4=32$; $\chi=2$

5-vacity: $m_0=12$, $m_1=60$, $m_2=120$, $m_3=240$, $m_4=192$, $m_5=64$; $\chi=0$

6-cavity: $m_0=14$, $m_1=84$, $m_2=280$, $m_3=560$, $m_4=672$, $m_5=448$, $m_6=128$; $\chi=2$

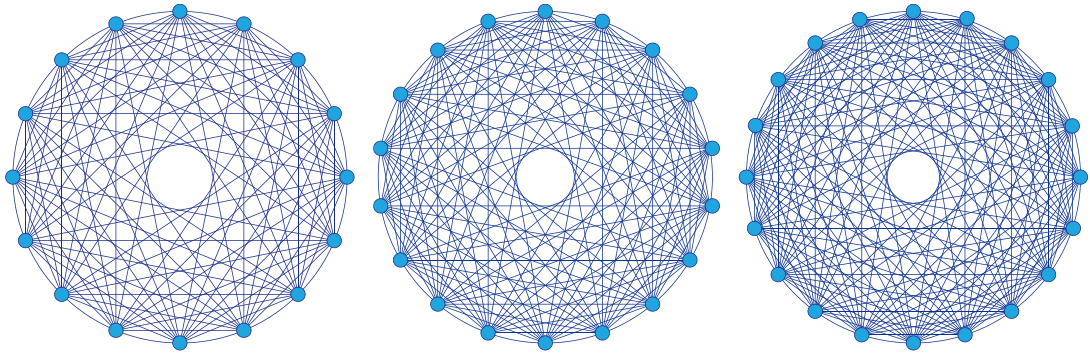

*7-cavity:* $m_0=16$, $m_1=112$, $m_2=448$, $m_3=1120$, $m_4=1792$, $m_5=1792$, $m_6=1024$, $m_7=256$; $\chi=0$

8-cavity: $m_0=18$, $m_1=144$, $m_2=672$, $m_3=2016$, $m_4=4032$, $m_5=5376$, $m_6=4608$, $m_7=2304$, $m_8=512$; $\chi=2$

9-cavity: $m_0=20$, $m_1=180$, $m_2=960$, $m_3=560$, $m_4=3360$, $m_5=8064$, $m_6=13440$, $m_7=15360$, $m_8=11520$, $m_9=1024$; $\chi=0$

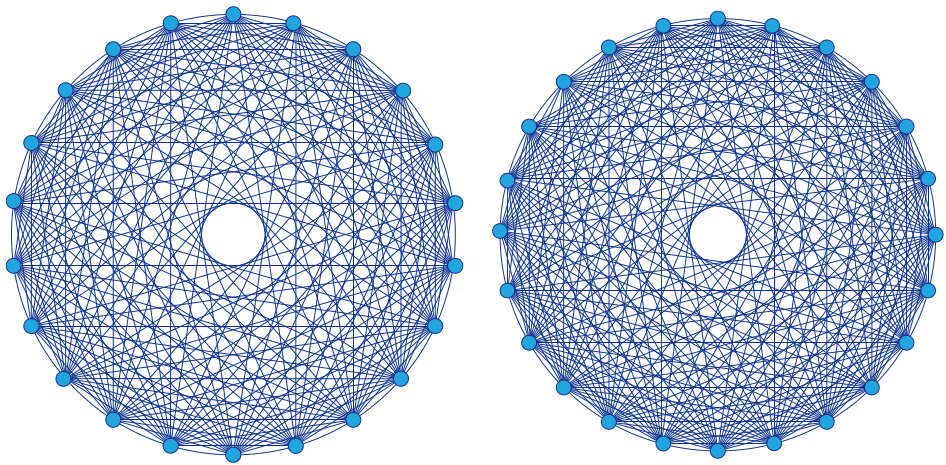

10-cavity: $m_0=22$, $m_1=220$, $m_2=1320$, $m_3=5280$, $m_4=14784$, $m_5=29568$, $m_6=42240$, $m_7=42240$, $m_8=28160$, $m_9=11264$, $m_{10}=2048$; $\chi=2$

11-cavity: $m_0=24$, $m_1=264$, $m_2=1760$, $m_3=7920$, $m_4=25344$, $m_5=59136$, $m_6=101376$, $m_7=125720$, $m_8=112640$, $m_9=67584$, $m_{10}=24576$, $m_{11}=4096$; $\chi=0$

**Figure SI-1.** Smallest cavities of order 1 to order 11.

**Supplementary Note 3. Cliques and Cavities in C. elegans Network**

For a dataset of C. elegans with 297 neurons and 2148 synapses[21], its cliques, ranks and the number of cavities are all obtained by using an available algorithm[9] and the proposed algorithm, with results summarized in Table SI-3.

**Table SI-3.** C. elegans Network

| Clique | $m_0$ | $m_1$ | $m_2$ | $m_3$ | $m_4$ | $m_5$ | $m_6$ | $m_7$ | $m_8$ |
|---|---|---|---|---|---|---|---|---|---|
|  | 297 | 2148 | 3241 | 2010 | 801 | 240 | 40 | 2 | 0 |
| Rank | $r_0$ | $r_1$ | $r_2$ | $r_3$ | $r_4$ | $r_5$ | $r_6$ | $r_7$ | $r_8$ |
|  | 0 | 296 | 1713 | 1407 | 599 | 202 | 38 | 2 | 0 |
| Betti number | $\beta_0$ | $\beta_1$ | $\beta_2$ | $\beta_3$ | $\beta_4$ | $\beta_5$ | $\beta_6$ | $\beta_7$ | $\beta_8$ |
|  | 1 | 139 | 121 | 4 | 0 | 0 | 0 | 0 | 0 |

Based on the data in Table SI-3, using the 0-1 programming, it is possible to find 4 different 3-cavities. The details are given below.

The first 3-cavity with 8 nodes (85,13,3,164,163,119,118,158) is surrounded by the following 16 3-cliques:
(85,13,3,164)  (13,3,164,163)  (3,164,163,119)  <u>(164,163,119,118)</u>
(163,119,118,158)  (119,118,158,85)  (118,158,85,13)  (158,85,13,3)
(85,3,164,119)  (119,158,85,3)  (3,163,119,158)  (158,13,3,163)
(163,118,158,13)  (13,164,163,118)  (118,85,13,164)  (164,119,118,85)

The second 3-cavity with 11 nodes (163, 3, 162, 119, 154, 167, 118, 227, 85, 13, 164) is surrounded by the following 28 3-cliques:
(163,3,162,119)  (3,162,119,154)  (162,119,154,118)  (119,154,118,167)
(154,118,167,13)  (118,167,13,227)  (167,13,227,3)  (13,227,3,85)
(227,3,85,119)  (3,85,119,164)  (85,119,164,118)  (119,164,118,163)
(118,163,119,162)  (162,118,163,13)  (13,162,118,154)  (154,13,162,3)

(3,154,13,167)   (167,3,154,119)   (119,167,3,227)   (227,119,167,118)
(118,227,119,85)   (85,118,227,13)   (13,85,118,164)   (164,13,85,3)
(3,164,13,163)   (163,3,164,119)   (13,163,118,164)   (163,3,162,13)

The third 3-cavity with 8 nodes (171,13,3,195,185,119,118,173) is surrounded by the following 16 3-cliques:
(171,13,3,195)   (13,3,195,185)   (3,195,185,119)   (195,185,119,118)
(185,119,118,173)   (119,118,173,171)   (118,173,171,13)   (173,171,13,3)
(171,3,195,119)   (119,173,171,3)   (3,185,119,173)   (173,13,3,185)
(185,118,173,13)   (13,195,185,118)   (118,171,13,195)   (195,119,118,171)

The fourth 3-cavity with 8 nodes (173,13,3,227,195,119,118,185) is surrounded by the following 16 3-cliques:
(173,13,3,227)   (13,3,227,195)   (3,227,195,119)   (227,195,119,118)
(195,119,118,185)   (119,118,185,173)   (118,185,173,13)   (185,173,13,3)
(173,3,227,119)   (119,185,173,3)   (3,195,119,185)   (185,13,3,195)
(195,118,185,13)   (13,227,195,118)   (118,173,13,227)   (227,119,118,173)

For 2-cavities, only those with eight nodes are listed here. A total of 4 have been found, which are divided into two types, as shown in Figure SI-2.

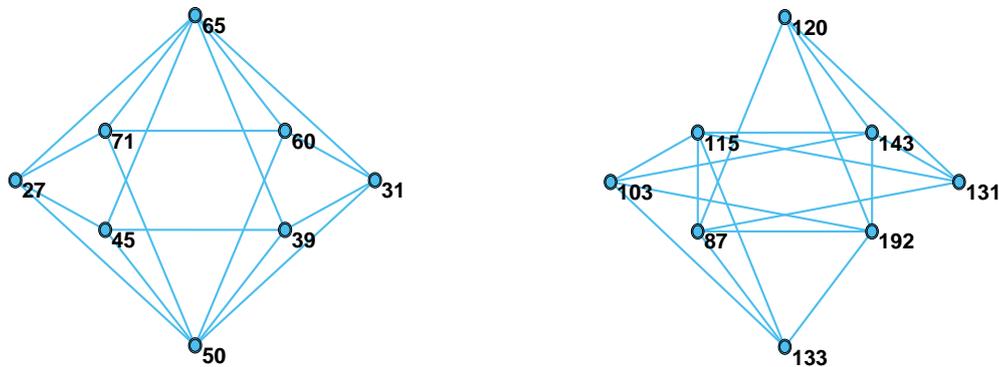

**Figure SI-2.** Two types of 2-cavities.

The first type of 2-cavity with 8 nodes (65, 31, 39, 45, 27, 71, 60, 50) is surrounded by the following 12 2-cliques:
(65,31,39)   (65,39,45)   (65,45,27)   (65,27,71)   (65,71,60)   (65,60,31)
(50,31,39)   (50,39,45)   (50,45,27)   (50,27,71)   (50,71,60)   (50,60,31)

The second type of 2-cavity with 8 nodes (120, 131, 143, 192, 87, 115, 103, 133) is surrounded by the following 12 2-cliques:
(120,131,143)   (120,131,87)   (120,192,87)   (120,143,192)
(103,143,192)   (103,115,143)   (103,115,133)   (103,133,192)
(87,133,192)   (87,115,133)   (87,115,131)   (115,131,143)